\documentclass[conference]{IEEEtran}
\IEEEoverridecommandlockouts
\usepackage{multirow}
\usepackage{multicol}
\usepackage{cite}
\usepackage{amsmath,amssymb,amsfonts}
\usepackage{mathtools}

\usepackage{algorithmic}
\usepackage{graphicx}
\usepackage{subcaption}
\usepackage{textcomp}
\usepackage{xcolor}
\def\BibTeX{{\rm B\kern-.05em{\sc i\kern-.025em b}\kern-.08em
    T\kern-.1667em\lower.7ex\hbox{E}\kern-.125emX}}
\begin{document}

\title{Analyzing the Effect of  Multi-task Learning for Biomedical Named Entity Recognition \\
\author{\IEEEauthorblockN{Arda Akdemir}
\IEEEauthorblockA{
\textit{University of Tokyo}\\
Tokyo, Japan \\
aakdemir@hgc.jp}
\and
\IEEEauthorblockN{Tetsuo Shibuya}
\IEEEauthorblockA{
\textit{University of Tokyo}\\
Tokyo, Japan \\
tshibuya@hgc.jp}}
}

\maketitle
\begin{abstract}
Developing high-performing systems for detecting biomedical named entities has major implications. State-of-the-art deep-learning based solutions for entity recognition often require large annotated datasets, which is not available in the biomedical domain. Transfer learning and multi-task learning have been shown to improve performance for low-resource domains. However, the applications of these methods are relatively scarce in the biomedical domain, and a theoretical understanding of why these methods improve the performance is lacking. In this study, we performed an extensive analysis to understand the transferability between different biomedical entity datasets. We found useful measures to predict transferability between these datasets. Besides, we propose combining transfer learning and multi-task learning to improve the performance of biomedical named entity recognition systems, which is not applied before to the best of our knowledge.

\end{abstract}

\section{Introduction}

Publicly available biomedical articles keep increasing rapidly. Automated systems that utilize biomedical text mining methods are necessary to be able to handle this large amount of data with minimal manual effort. An important first step of any biomedical text mining method is the detection and classification of biomedical entities such as disease, drug or chemical mentions in biomedical articles. This task is referred to as biomedical named entity recognition (BioNER).

BioNER  has seen remarkable progress with the advents in machine learning and deep learning methods~\cite{lee2019biobert}. These methods require labeled datasets and benefit from increasing the amount of labeled examples. Artificial neural networks form the core of almost all state-of-the-art bioNER systems. The main drawback of these methods is that the networks must be trained from scratch for each dataset, separately. Even though recent progress in BioNER is promising, the overall performance is significantly lower than general domain NER. This is mainly due to the scarcity and sub-optimal utilization of the labeled datasets in the biomedical domain.

Transfer learning is a training paradigm that mitigates the above mentioned issues of current approaches. It attempts to utilize the information obtained from a source task to improve the performance on a target task. Transfer learning is shown to be especially useful when the size of the labeled data is limited for the target task~\cite{ruder2019neural}, making BioNER a very suitable target task. Multi-task learning is a special case of transfer learning where multiple tasks are learned simultaneously~\cite{caruana1997multitask}. In this context, this corresponds to learning multiple biomedical named entity datasets using a single neural network.

Recently, the seminal work of Devlin et al.~\cite{devlin2019bert}, i.e. the BERT model, enabled progress in various NLP tasks, including NER. BERT uses self-supervised learning which relieves the need for having labeled examples to train the neural network. Lee et al.~\cite{lee2019biobert} proposed BioBERT, which is the BERT model pretrained on a large unlabeled biomedical dataset. They finetuned the BioBERT model on labeled datasets using supervised learning and obtained improvements on several downstream biomedical NLP tasks. Yet, BioBERT has not been applied before in the context of multi-task learning, to the best of our knowledge. This motivated us to use BioBERT as the shared network across all biomedical datasets. We claim that sharing information across datasets help improve the overall performance as the representations obtained on one biomedical dataset is relevant to others, even though the annotated entities are different.

Multi-task learning is also used recently to improve the performance on BioNER datasets~\cite{wang2019cross,crichton2017neural}. Yet, the analysis of where the improvements come from is limited and the effect of transfer learning is unclear. Thus, there is a lack of theoretical understanding of when and why transfer learning and multi-task learning bring improvements.

In this study, we analyze the effect of multi-task learning for biomedical named entity recognition. To this end, we experimented on seven BioNER benchmark datasets and analyzed the effect of multi-task learning by using ten different measures. We evaluate the usefulness of these measures with three different metrics. Besides, we propose combining transfer learning and multi-task learning for BioNER which is not employed before to the best of our knowledge. The main contributions of this study are as follows:

\begin{itemize}
    \item We analyzed the effect of multi-task learning by using various dataset characteristics, similarity measures, and evaluation metrics. This allowed us to find useful features and measures to predict transferability, which we believe can help reduce the required manual effort for finding useful auxiliary datasets in future.
    \item We propose a novel BioBERT-CRF based single task model that outperformed the state-of-the-art on four datasets.
    \item We incorporated multi-task learning to the proposed neural network which brought performance improvements on six out of eight cases.
\end{itemize}

\section{Related Work}
Multi-task learning and transfer learning have both been applied for BioNER before. Lee et al.~\cite{lee2019biobert} obtained state-of-the-art results on several biomedical entity recognition datasets by using the BERT language model pretrained on the Biomedical abstracts and papers.  Crichton et al.~\cite{crichton2017neural} is the first work that attempts to apply multi-task learning for detecting biomedical named entities. They used pretrained biomedical word embeddings and used a CNN based neural network to detect the named entities. They also use Viterbi decoding for the tag transition probabilities. They only calculate binary probabilities for tag transitions (appears or does not appear inside the training dataset), whereas we use a CRF layer with trainable parameters to output a score for each possible transition. Wang et al.~\cite{wang2019cross} proposed a BiLSTM-CRF based neural network where  character and word embeddings are shared by all dataset-specific components. Unlike  Crichton et al., who propose sharing a convolutional layer, they propose sharing only word/character embeddings across different datasets. Mehmood et al.~\cite{mehmood2019multi} also analyzed the effect of multi-task learning for BioNER datasets. They propose using stack LSTMs for multi-task learning of BioNER datasets. Their analysis of multi-task learning focus on \textit{architectural variations}, whereas we focus on analyzing \textit{dataset characteristics} for transferability.
Yoon et al.~\cite{yoon2019collabonet} proposed using an expert BiLSTM-CRF based network for each dataset, separately. During training, they use the output of all other trained models as an additional input to the neural network. They claim that the output of each expert model helps prevent mislabelings caused by \textit{polysemy}.

Even though both transfer learning and multi-task learning have already been used for BioNER, the analysis of the effect of using these methods is highly limited. A similar analysis to ours is done before on general domain NLP datasets. Bingel et al.~\cite{bingel2017identifying} analyzed the task relations for multi-task learning on ten NLP tasks. Alonzo et al.~\cite{alonso2016multitask} analyzed the effect of multi-task learning on several sequence labeling tasks. Different from these previous studies,w we focus on the datasets from the biomedical domain. These datasets have unique characteristics and we claim that findings on the general domain datasets might not be directly applicable to them.

\section{Datasets}

In our experiments we used seven biomedical entity datasets. BC5CDR contains both disease and chemical entity mentions. We follow the same convention with previous work to have two identical copies of the dataset that are only annotated for each entity type (BC5CDR-chem and BC5CDR-disease). The details about the datasets are given in Table~\ref{dataset_stats}.

\begin{table}
    \centering
        \caption{Details about the biomedical entity datasets}
    \label{dataset_stats}
    \resizebox{0.45\textwidth}{!}{
    \begin{tabular}{c|c|c|c}\hline
&Entity type&Number of sentences&Entity/token ratio\\\hline
BC2GM~\cite{smith2008overview}&	Gene/Protein NE&12,573&0.043\\
BC4CHEMD~\cite{krallinger2015chemdner}&	Chemical NE&30,681&0.033\\
BC5CDR-chem~\cite{wei2012proceedings}&	Chemical NE&4,559&0.044\\
BC5CDR-disease~\cite{wei2012proceedings}&Disease NE&4,559&0.035\\
JNLPBA~\cite{collier2004jnlpba}&Gene/Protein NE&14,689&0.073\\
NCBI-disease~\cite{dougan2014ncbi}&Disease NE&5,423&0.038\\
 linnaeus~\cite{gerner2010linnaeus}& Species NE&11,934&0.008\\
s800~\cite{pafilis2013species}&Species NE&5,732&0.017\\\hline
    \end{tabular}}

\end{table}

\section{Dataset measures and evaluation metrics for analyzing transferability}
\label{measures}

In order to understand under which conditions multi-task learning brings more benefits, we used a total of 12 dataset  measures. To take into account the asymmetric characteristics of MTL gains we included directed similarity measures. 

\subsection{Dataset measures}

\textbf{Shared Vocabulary} (vocab) (directed): Rate of the target vocabulary words found inside the auxiliary dataset vocabulary~\cite{schroder2020estimating}. This is a directed metric as the vocabulary sizes differ among datasets.
$$
s = \dfrac{|V_t\cap V_a |}{|V_t|}
$$
where $s$ is the shared vocabulary ratio for a target dataset $t$ and an auxiliary dataset $a$, where $V$ represents the vocabulary.
\textbf{Topic Distribution similarity} (topic) (undirected): Following Ruder et al.~\cite{ruder2017learning}, we used LDA topic modeling~\cite{blei2003latent} to represent each dataset as $n$ dimensional vectors, where each index represents a topic. We calculated the cosine similarity between all dataset pairs:

$$
t_{i,j} = t_{j,i} = \dfrac{dot(v^t_i,v^t_j)}{\|v^t_i\|\|v^t_j\|}
$$
where $v^t_i$ represents the $n$ dimensional topic vector of the $i^{th}$ dataset, and $t_{i,j} = t_{j,i}$ is the topic based similarity score.

\textbf{Embedding similarity} (bert) (undirected): Word embedding based distance is frequenly used to measure similarity between datasets~\cite{ruder2017learning}. Recently,Vu et al.\cite{vu2020exploring} used 
BERT-based representations to measure similarity between datasets. Similarly, we leverage the BioBERT model to obtain the vector representations of all sentences of a dataset. Then we average over all sentences to get the BioBERT-based dataset representations:
\begin{align*}
    v^b_i &= \dfrac{1}{|D_i|} \sum_{k=1}^{|D_i|} BioBERT(D_i(k))\\
    b_{i,j}&=b_{j,i} =  \dfrac{dot(v^b_i,v^b_j)}{\|v^b_i\|\|v^b_j\|}
\end{align*}
where $D_i(k)$ is the $k^{th}$ sentence of the $i^{th}$ dataset, $v^b_i$ is the BioBERT representation of $i^{th}$ dataset, and $ b_{i,j}$ is the embedding-based similarity.

\textbf{Cooccurring entity ratio} (cooccur, directed): Rate of the target entity tokens found and annotated inside the auxiliary dataset. Schroder et al.~\cite{schroder2020estimating} show that the rate of cooccurring entities is a good predictor for MTL gains. We follow them to include this value to predict transferability.

We first used each of the above measures to rank all auxiliary datasets for a specific target dataset, separately. Then, we considered pairwise combinations of these four measures. In total this gave us 10 different scores about the transferability for $(T,A)$ pair where $T$ is the target dataset, and $A$ is the auxiliary task. Note that these scores are different for $(A,T)$ and $(T,A)$ as we include directed similarity measures.

\subsection{Evaluation Metrics.}

To evaluate the usefulness of each measure, we used three different metrics. This analysis provide useful hints about which measures to use in future to determine the auxiliary datasets/tasks for multi-task learning. 

We use the metrics below and the final F1 scores obtained on the target tasks, to evaluate the quality of the rankings calculated by each measure. Ideally, a ranking  is perfect if the ordering completely overlaps with the ordering of the auxiliary datasets if we sort them in descending order based on the F1-score on the target dataset.

\textbf{Normalized Discounted Cumulative Gain} (NDCG): Following \cite{vu2020exploring}  we used this metric to evaluate the quality of the rankings. Normalized Discounted Cumulative Gain is a frequently used measure to evaluate the goodness of a ranking\cite{jarvelin2002cumulated} (e.g., search engines). For $n$ auxiliary tasks NDCG is calculated as follows:
\begin{align*}
DCG_n(R) &=\sum_{i=1}^{n}\dfrac{2^{rel_i}-1}{log_2(i+1)} \\
  NDCG^j_{n} &= \dfrac{DCG_n(R^j)}{DCG_n(\textbf{R})}\\
\end{align*}
where $rel_i$ is the $i^{th}$ auxiliary task according to a ranking $R$, $R^j$ is the predicted ranking of the auxiliary datasets according to the $j^{th}$ similarity measure, and $\textbf{R}$ is the ideal ranking according to the target dataset performance. When there is a perfect alignment between $R^j$ and $\textbf{R}$, NDCG score is 1. Higher NDCG scores point to a better ranking.

\textbf{Average rank of the best auxiliary dataset} $(\rho)$: Average of the ranks  of the best auxiliary datasets for each target datasets, according to the rankings calculated by using a similarity measure:\begin{align*}
    \rho_j = \dfrac{1}{n} \sum_{i=1}^{n} R_j(a_i)
\end{align*}where $R_j(a_i)$ is the rank of the best auxiliary task ($a_i$) for the $i^{th}$ target task. This measures how far we are from the ideal auxiliary dataset using a specific similarity measure. If the similarity measure finds the best auxiliary dataset in all cases, this score will be equal to 1. Lower scores point to a more accurate ranking.

\textbf{Average rank of the best auxiliary prediction} $(\sigma)$: Average of the ranks of the top auxiliary auxiliary predictions according to the ranking by the MTL gains:\begin{align*}
    \sigma_j = \dfrac{1}{n} \sum_{i=1}^{n} \textbf{R}(a^j_i)
\end{align*}where $\textbf{R}(a^j_i)$ is the rank of the best auxiliary task prediction of $j^{th}$ similarity measure for the $i^{th}$ target dataset according to the ideal ranking $\textbf{R}$.
This measures the real rank of the top prediction of a similarity measure. Similarly in the ideal situation, a perfect similarity measure will always return the best auxiliary dataset as the top prediction and this score will be equal to 1, and lower scores point to a more accurate ranking.

\section{Experiment Details}

All our experiments are conducted using a single-thread V100 GPU. We use the PyTorch library to implement the proposed neural networks.

For the initial single-task learning experiments, we applied early stopping on the validation split to determine the number of steps on each datasets.In order to have a fair comparison, we train the single-task and multi-task models for equal number of iterations on each target dataset. We keep the number of iterations on the auxiliary and target datasets the same during the multi-task learning experiments. So, if a single-task model is trained for $X$ steps on a target dataset, the multi-task counterpart is  trained for $X$ steps on the target and $X$ steps on the auxiliary dataset.

\begin{table*}[ht]
\centering
\caption[NER2]{Results for single task learning of Biomedical NER. } 
\label{table:stlres}

\begin{tabular}{c| c c c|c c c| ccc}\hline

&  \multicolumn{3}{c}{BioBERT baseline} &  \multicolumn{3}{c}{BioBERT+CRF} & \multicolumn{3}{c}{SOTA}\\\hline
DATASET	&PRE	&REC&	F-1& PRE	&REC&	F-1&PRE	&REC&	F-1\\\hline
BC2GM &82.19 &	83.93 &	83.05&82.86 &	84.61 &	83.73&	84.32&	85.12 &	\textbf{84.72} \\
BC4CHEMD&	90.12 &	90.74 & 	90.43 &	91.74&	90.69&	91.21 &	92.80 &	91.92 &	\textbf{92.36} \\
BC5CDR-chem&	93.08 &	93.87	& 93.47 &93.29&	93.72	& \textbf{93.76}&	93.68 &	93.26 &	93.47 \\
BC5CDR-disease &	83.78 &	86.87 &	85.3  &85.56 &	86.69&	86.12&	86.47 &	87.84 &	\textbf{87.15}\\
JNLPBA&70.9 &	82.65 &	76.32&	71.79	&85.11	&\textbf{78.71}&	72.24 &	83.56 &	77.49 \\
NCBI-disease &	86.43 &	88.23 & 	87.32 &	87.21	&90.21&	88.89 &	88.22 &	91.25 &	\textbf{89.71}\\
linnaeus &	91.36 & 	87.97 & 	89.63	&91.72&	85.94&	\textbf{89.16} &	90.77 &	85.83 &	88.24\\
s800& 	70.11 & 	76.14 & 	73.0 &	72.0	&80.44&	\textbf{75.99} &	 72.80 &	75.36 &	74.06\\\hline

\end{tabular}

\end{table*}

We conducted single-task experiments on eight different datasets. We also conducted multi-task related experiments  for all pairs of datasets (36 pairs). In total, we ran experiments on 44 different settings.  

\section{Results}

\subsection{Single Task Learning Results}
As the baseline models, we trained BioBERT based single-task learning models for each dataset, separately. Next, we incorporated a CRF layer to this baseline model. This improved the results obtained on each dataset significantly. Table~\ref{table:stlres} shows the results for STL. The results show that incorporating CRF layer improves the performance across all datasets. These results also show that our proposed BioBERT+CRF single task model achieves SOTA results for four out of eight biomedical datasets. For these experiments, we follow Lee et al.~\cite{lee2019biobert} to combine the validation and training datasets to train our models, in order to have a fair comparison.

\subsection{Multi Task Learning Results}

Next, we give the results for the multi-task learning models. For these experiments, we did not combine the training and validation sets. Validation sets are used to determine when to stop the training. Yet, we still report the SOTA results on the same table for reference.
In total we used eight datasets and we trained multi-task learning models for all pairs. We used a flat architecture that uses the pretrained BioBERT model as the shared layer. Results on Table~\ref{table:nerreswithall} show that incorporating multi-task learning improves performance on all eight datasets tested. This confirms our claim that multi-task learning helps transfer useful information across biomedical datasets with different entity labels. Besides, we observed that multi-task learning models outperformed the state-of-the-art on four datasets. 

\begin{figure}
\centering
  \includegraphics[width=0.45\linewidth]{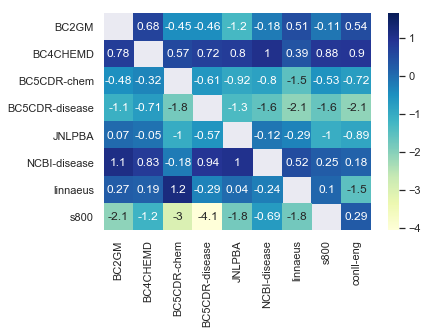}
    \includegraphics[width=0.45\linewidth]{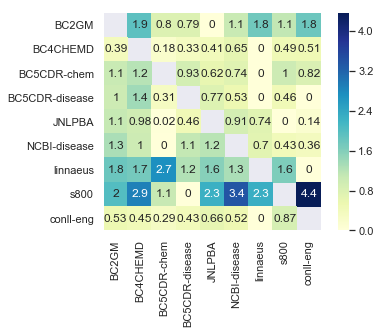}
    \caption{Heatmaps showing the relative performance of multi-task learning over single-task learning(left) and the relative performance of using different auxiliary dataset (right). For each row on the right figure, the auxiiliary dataset (column) that brought the least improvement is denoted with zero.}
    \label{fig:hists}
\end{figure}

Figure~\ref{fig:hists} shows the relative performance difference by incorporating multi-task learning. The horizontal axis on each heatmap denotes the target datasets. We see that some datasets benefit more from multi-task learning than others (e.g., BC4CHEMD performance increased significantly by transferring information from all datasets). We see that BC5CDR-disease and  BC5CDR-chem datasets benefited the least from multi-task learning. We only observed performance degradation by incorporating multi-task learning in 4 out of 56 experiments. Another interesting observation is that, the benefits are not symmetric. For example, transferring information from s800 dataset to the BC5CDR-disease dataset degraded the performance by 0.15 F-1 Score. However, using BC5CDR-disease as the auxiliary dataset for the s800 dataset gave the second highest improvement on the s800 target dataset. This motivated us to consider directional similarity measures (in addition to undirected measures) to predict transferability between datasets.

\textbf{Effect of dataset size for multi-task learning.}
We analyzed the effect of the dataset sizes to the MTL gains.
Table~\ref{table:feat_corr} shows the correlation coefficients between datasets sizes and MTL gains. We see that for both correlation methods, target dataset size seem to correlate well with MTL gains. Specifically, \textit{larger datasets benefit more from multi-task learning}. The correlation between auxiliary dataset sizes and mtl gains is less pronounced. We see a slight negative correlation between the auxiliary dataset sizes and mtl gains. However, the correlations are not significant enough to conclude that small auxiliary datasets would bring more mtl gains.

\begin{table}[ht]
\centering
\caption[corr]{Correlation between MTL gains and dataset characteristics.} 
\label{table:feat_corr}

\resizebox{0.45\textwidth}{!}{
\begin{tabular}{c| c |c|c|c}\hline
& pearson  &p  & spearman  &p \\\hline
Target size & 0.477&0.2324&0.443&0.2715\\
Auxiliary size &-0.301&0.4689&-0.383&0.3487\\\hline
Target entity/token ratio &-0.382&0.35&-0.5&0.207\\
Auxiliary entity/token ratio &0.698&0.054&0.833&0.01\\\hline
\end{tabular}}

\end{table}

\begin{figure*}
    \begin{subfigure}{0.25\textwidth}
  \centering
    \includegraphics[width=.9\linewidth]{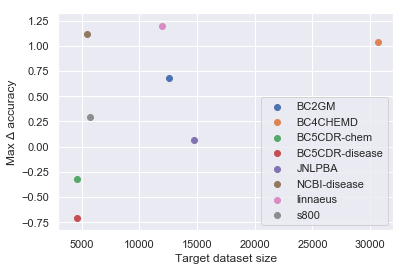}
    \caption{}
    \label{fig:mtl_size_1}
\end{subfigure}%
\begin{subfigure}{0.25\textwidth}
  \centering
    \includegraphics[width=.9\linewidth]{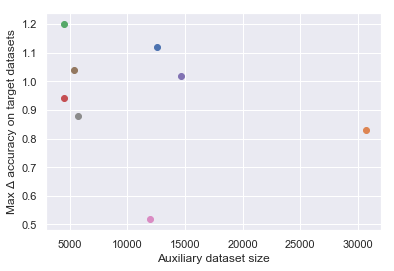}
    \caption{}
    \label{fig:mtl_size_2}
\end{subfigure}%
  \begin{subfigure}{0.25\textwidth}
    \includegraphics[width=.9\linewidth]{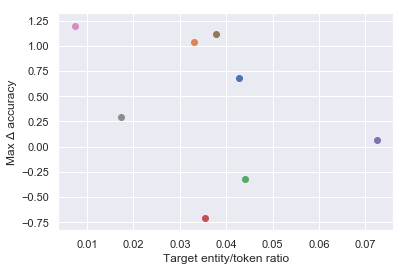}
    \caption{}
    \label{fig:target_entrat}

\end{subfigure}%
 \begin{subfigure}{0.25\textwidth}
    \includegraphics[width=.9\linewidth]{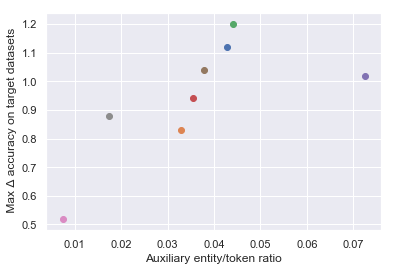}
    \caption{}
    \label{fig:aux_entrat}
    \end{subfigure}%
   
\caption{Plots showing the relation between dataset characteristics and change in accuracy. (a-b) show the relation between dataset sizes and change in target task accuracy. (c,d) show the relation between the target accuracy change and the entity/token ratios of auxiliary and target datasets, respectively. The correlation of the auxiliary entity/token ratio and the correlation of the target dataset size with the change in accuracy is stronger.}
\end{figure*}

\textbf{Effect of entity/token ratio for multi-task learning.}
We analyzed the effect of the entity/ sizes to the MTL gains. Previous work on analyzing the effect of multi-task learning also utilized this ratio ratio~\cite{alonso2016multitask,bingel2017identifying}. Bingel et al.~\cite{bingel2017identifying} report that both the target token/type ratio and the auxiliary token/type ratio are not among the best predictors for multi-task learning gains (16th and 20th strongest predictors out of 42 features). They found out that the target token/type ratio is negatively correlated with the mtl gains, whereas the auxiliary token/type ratio is positively correlated. Table~\ref{dataset_stats} shows the entity ratio for all biomedical entity recognition datasets we have used. We see that there is a significant variance between datasets for this entity/token ratio. It varies from $0.8\%$ (linnaeus) to $7.3\%$ (JNLPA). 

Table~\ref{table:feat_corr} shows the correlation coefficients between entity/token ratios and MTL gains. We see that the target entity/token ratio is negatively correlated (-0.382 and -0.5 for pearson and spearman correlations, respectively) whereas auxiliary entity/token ratio is positively correlated (0.698 and 0.833 for pearson and spearman correlations, respectively). In other words, \textit{multi-task gains are more likely to occur when the target dataset entities are sparse and the auxiliary dataset entities are abundant.} The correlation for the auxiliary entity/token ratio is stronger and seems to be a better predictor for multi-task gains in this setting.
We speculate that the abundance of auxiliary entities may help capturing important representations of the biomedical entities without simply memorizing them.



%
\begin{table*}[ht]
\centering
\caption[NER3]{Comparison of all multi-task learning experiments with the single-task learning counterparts and the current state-of-the-art. All results are given in percentage  (\%) F-1.  } 
\label{table:nerreswithall}
\resizebox{\textwidth}{!}{
\begin{tabular}{c| c| c| cccc cccc c c}\hline

&STL&SOTA&BC2GM&BC4CHEMD&BC5CDR-chem&BC5CDR-disease&JNLPBA&NCBI-disease&linnaeus&s800&conll-eng&All\\\hline
BC2GM&82.31&84.72&-&82.99&81.86&81.85&81.06&82.13&82.82&82.2&82.85&80.76\\
BC4CHEMD&88.43&92.36&89.21&-&89.0&89.15&89.23&89.47&88.82&89.31&89.33&87.71\\
BC5CDR-chem&92.9&93.47&92.42&92.58&-&92.29&91.98&92.1&91.36&92.37&92.18&91.3\\
BC5CDR-disease&85.57&87.15&84.49&84.86&83.78&-&84.24&84.0&83.47&83.93&83.47&84.49\\
JNLPBA&77.81&77.49&\textbf{77.88}&77.76&76.8&77.24&-&77.69&77.52&76.78&76.92&77.56\\
NCBI-disease&86.98&89.71&88.1&87.81&86.8&87.92&88.0&-&87.5&87.23&87.16&88.09\\
linnaeus&87.95&88.24&88.22&88.14&\textbf{89.15}&87.66&87.99&87.71&-&88.05&86.41&87.03\\
s800&75.33&74.06&73.27&74.17&72.34&71.26&73.54&74.64&73.55&-&\textbf{75.62}&74.15\\
conll-eng&-&93.09&88.4&88.32&88.16&88.3&88.53&88.39&87.87&88.74&-&-\\\hline
\end{tabular}}

\end{table*}

\subsection{Learning All Datasets at Once}  Next, we attempted to train a single multi-task learning model to learn all datasets at once. This methodology is different than training a single model on the concatenation of all datasets, as the architecture contains task-specific sequence labeling layers for each dataset. The rightmost column of Table~\ref{table:nerreswithall} shows the results of this experiment. We observe that learning all datasets at once only brought improvement over single-task learning for the NCBI-disease dataset. We observe that except for BC2GM, BC4CHEMD, and BC5CDR-chem datasets, learning all datasets at once performed better than some  pairwise multi-task learning setting. Yet, for all target datasets, pairwise multi-task learning outperformed learning all datasets at once for at least one setting. These results strongly suggest that the model struggles to learn good representations on the common layer when it is exposed to all datasets at once. The results also show that naively exposing the neural network to more data is not guaranteed to bring performance improvements.

\subsection{Analysis of Transferring Information between Datasets}
Finally, we analyze the transferability between datasets. The aim of this analysis is to understand when and why transfer learning helps improve performance on a target dataset. For this reason, we also included Conll-2003 English Named Entity Recognition dataset as an out-of-domain dataset. The results are given in Table~\ref{table:nerreswithall}. Surprisingly, we did not observe that transferring information from the Conll dataset (out-of-domain) performs worse than transferring information from all other biomedical entity datasets (in-domain). Only for the linnaeus and BC5CDR-disease datasets, we observe that using Conll as the auxiliary dataset has the lowest performance.

Right heatmap of Fig.\ref{fig:hists} shows the relative performance on each target dataset. For each row (target dataset) the lowest scoring auxiliary dataset has the value zero. We see that the benefits of multi-task learning is not symmetric. This motivated us to incorporate directed similarity measures to analyze the multi-task gains (explained in section \ref{measures}).

\textbf{Multi-task learning in grouped datasets.} The seven biomedical datasets contained four types of entities (gene/protein, chemical, disease and species). JNLPBA and BC2GM contain gene/protein mentions. For the gene/protein group, we see that in-group transfer is very useful for JNLPBA but very harmful for BC2GM(lowest performance for BC2GM, and highest improvement for JNLPBA). BC4CHEMD and BC5CDR-chem contain chemical entity mentions. For the chemical entity group, we see that in-group transfer gives the second lowest performance for BC4CHEMD (transferring from BC5CDR-chem to BC4CHEMD), but highest performance for BC5CDR-chem.  For disease mention datasets (NCBI-disease and BC5CDR-disease) in-group transfer brought performance degradation for BC5CDR-disease (-1.6 F-1 Score over single-task learning), whereas it brought the third highest improvement +0.94 F-1 score over single task learning) for the NCBI-disease dataset.
For s800 and linnaeus (species group) datasets, we see that in-group transfer gives the $4th$ highest performance out of 9 datasets in both cases. The non-symmetry between the mtl gains in these pairwise setting is interesting and a detailed analysis might yield important findings.

\begin{table}[ht]
\centering
\caption{Evaluation of different similarity measures for predicting MTL gains.} 
\label{table:sim_eval}
\resizebox{0.45\textwidth}{!}{

\begin{tabular}{c| c|c|c}\hline
&NDCG Scores&Best auxiliary dataset rank&Rank of best auxiliary prediction\\\hline
random&0.688&4.5&4.5\\
topic&0.897&3.333&3.778\\
vocab&0.906&3.889&3.667\\
cooccur&0.918&3.778&3.889\\
bert&0.886&4.444&4.667\\
topic\_vocab&0.896&3.778&4.333\\
topic\_cooccur&0.905&3.444&3.778\\
topic\_bert&0.895&3.556&4.333\\
vocab\_cooccur&0.895&3.889&4.444\\
vocab\_bert&0.906&3.778&3.556\\
cooccur\_bert&0.895&3.889&4.444\\\hline
\end{tabular}}

\end{table}




Next, we used the similarity measures and evaluated the predictability of the MTL gains. Table~\ref{table:sim_eval} shows the results for measuring the goodness of all similarity measures. For NDCG, higher scores are better, whereas for the other two lower scores indicate a more accurate ranking. The first row (random) shows the values for randomly selecting the auxiliary datasets. NDCG score for random is generated by randomly scoring auxiliary datasets for 10,000 iterations. For all similarity measures, the NDCG score is significantly higher (at least 0.207) than selecting the auxiliary dataset randomly.

\textbf{Observations.}The first observation is that, \textit{we see a strong signal in the similarity measures we use}. Using any of the similarity measures to pick an auxiliary dataset is better than making a random selection. We observe that in our setting, \textit{cooccur} produced the best ranking according to the NDCG score. Similarity measures using \textit{topic} (topic alone or its combination with other similarity measures) occupies the top $4$ best average ranking for the best auxiliary dataset ($\rho$ metric). This result strongly suggests that utilizing the topic-based similarity information is very useful for finding beneficial auxiliary datasets. In other words, \textit{topic based similarity scores obtained using an LDA model is a good predictor for MTL gains}. 

Moreover, we observed that combining topic similarity with any other similarity measure always improved the $\rho$ score. Another critical observation is related to the BERT-based similarity. We see that BERT-based similarity performed the lowest according to all three evaluation methods. This result strongly suggests that using \textit{the BERT-based dataset vector similarities is not a good predictor for mtl gains}. We suspect that this might be caused by large number of entities that occur but are not annotated inside the auxiliary datasets with similar BioBERT vectors. These unlabeled entities might hinder the target dataset performance.

\section{Conclusion}
In this work, we proposed combining transfer learning and multi-task learning for BioNER, which is not done before to the best of our knowledge. The proposed method achieved state-of-the-art results on several biomedical named entity datasets. The main purpose of this study was to analyze and understand under which conditions transferring information from an auxiliary dataset helps improve performance on a target dataset. To this end, we used various dataset measures and evaluated their ability to predict the MTL gains using three different evaluation metrics. The analysis showed that the dataset measures contain strong signals about the benefits of multi-task learning.

\section{Future Work}
First, we plan to expand the analysis of predictability of MTL gains by including more dataset measures. Next, we plan to focus on data selection. We aim to focus on auxiliary datasets that brought the least MTL gains for a target dataset, and find subsets of these datasets that will bring higher MTL gains.
\bibliography{bioner}
\bibliographystyle{ieee_natbib}
\newpage

\appendix
\label{appendix}

\section{Implementation Details}
\label{app:imp}

All models are implemented using Python 3.7.2 programming language and the PyTorch library version 1. All the required source code and links to the relevant datasets to replicate the results are included inside the Supplemental Material. Please refer to the README.md document inside the Supplemental Material on how to run the codes and download the necessary materials.

Table~\ref{table:mtl_small} shows the results on the small datasets. We repeated each experiment five times and report the average score on each target dataset.

\begin{table*}[ht]
\centering
\caption[NER1]{F-1 scores for multi task learning of Biomedical NER datasets with 5 repeats with Conll.} 
\label{table:mtl_small}
\resizebox{\textwidth}{!}{
\begin{tabular}{c| c cccc cccc}\hline
&BC2GM&BC4CHEMD&BC5CDR-chem&BC5CDR-disease&JNLPBA&conll-eng&NCBI-disease&linnaeus&s800\\\hline
BC2GM&-&79.45&77.32&79.13&76.11&77.03&78.81&79.04&76.71\\
BC4CHEMD&85.7&-&84.05&84.26&84.75&86.03&84.73&85.01&85.28\\
BC5CDR-chem&91.76&93.09&-&92.15&91.93&92.41&91.87&92.19&92.37\\
BC5CDR-disease&81.46&82.0&82.3&-&81.63&82.6&81.25&81.5&81.53\\
JNLPBA&76.32&75.73&76.73&75.45&-&75.82&75.34&75.9&76.3\\
conll-eng&83.31&84.0&84.38&84.29&83.85&-&84.02&83.56&85.95\\
NCBI-disease&84.85&83.64&83.64&81.01&82.21&83.95&-&87.43&84.15\\
linnaeus&84.48&82.93&86.03&86.43&86.96&82.91&83.15&-&83.83\\
s800&78.01&74.29&75.71&80.85&77.14&79.1&80.0&78.87&-\\\hline

\end{tabular}}

\end{table*}

\end{document}